\documentclass[letterpaper, 10 pt, conference]{ieeeconf}  % Comment this line out if you need a4paper

\IEEEoverridecommandlockouts                              % This command is only needed if 
\usepackage{amsmath,amssymb,amsfonts}
\usepackage{algorithmic}
\usepackage{graphicx}
\usepackage{textcomp}
\usepackage{xcolor}
\usepackage{tikz}
\usepackage{soul}
\sethlcolor{yellow}
\usepackage{booktabs}
\usetikzlibrary{fit,backgrounds}
\usetikzlibrary{arrows.meta,positioning,calc}
\usetikzlibrary{positioning,fit,backgrounds,calc}
\usepackage{cite}
\usepackage[colorlinks=true,citecolor=blue,linkcolor=blue,urlcolor=blue]{hyperref}
\usepackage{pifont}
\def\BibTeX{{\rm B\kern-.05em{\sc i\kern-.025em b}\kern-.08em
    T\kern-.1667em\lower.7ex\hbox{E}\kern-.125emX}}

\title{\LARGE \bf
E-TIDE: Fast, Structure-Preserving Motion Forecasting from Event Sequences
}

\author{
Biswadeep Sen$^{1,2}$, Benoit R. Cottereau$^{2,3}$, Nicolas Cuperlier$^{2,4}$, Terence Sim$^{1,2}$ \\\\
\\
$^{1}$ National University of Singapore, Singapore \\
$^{2}$ IPAL CNRS IRL 2955, Singapore \\
$^{3}$ CerCo, CNRS UMR 5549, Universit\'{e} de Toulouse, France \\
$^{4}$ ETIS UMR8051, CY Cergy Paris Universit\'{e}, ENSEA, CNRS, Cergy, France
}

\begin{document}

\maketitle
\thispagestyle{empty}
\pagestyle{empty}

%%%%%%%%%%%%%%%%%%%%%%%%%%%%%%%%%%%%%%%%%%%%%%%%%%%%%%%%%%%%%%%%%%%%%%%%%%%%%%%%
\begin{abstract}
Event-based cameras capture visual information as asynchronous streams of per-pixel brightness changes, generating sparse, temporally precise data. Compared to conventional frame-based sensors, they offer significant advantages in capturing high-speed dynamics while consuming substantially less power. Predicting future event tensors from past observations is an important problem, enabling downstream tasks such as future semantic segmentation or object tracking without requiring access to future sensor measurements. While recent state-of-the-art approaches achieve strong performance, they often rely on computationally heavy backbones and, in some cases, large-scale pretraining, limiting their applicability in resource-constrained scenarios. In this work, we introduce E-TIDE, a lightweight, end-to-end trainable architecture for event-tensor prediction that is designed to operate efficiently without large-scale pretraining. Our approach employs the TIDE module (Temporal Interaction for Dynamic Events), motivated by efficient spatiotemporal interaction design for sparse event tensors, to capture temporal dependencies via large-kernel mixing and activity-aware gating while maintaining low computational complexity. Experiments on standard event-based datasets demonstrate that our method achieves competitive performance with significantly reduced model size and training requirements, making it well-suited for real-time deployment under tight latency and memory budgets.

\end{abstract}

% \begin{IEEEkeywords}
% Event cameras, event-based vision, spatiotemporal prediction, Deep Learning for Visual Perception, real-time robotic perception
% \end{IEEEkeywords}

%\section{First section}

\section{Introduction}
Robotic perception is inherently latency-critical: decisions must often be made under strict real-time constraints, before additional measurements arrive. This is especially true in fast motion, low light, and high dynamic range scenes, where conventional RGB cameras can degrade due to motion blur, exposure limitations, and bandwidth--latency trade-offs at high frame rates. Event cameras offer a compelling alternative. By reporting per-pixel brightness changes asynchronously, they produce sparse measurements with microsecond-level temporal precision, low latency, and high dynamic range \cite{gallego2020event}. These characteristics have enabled strong progress in event-based detection, tracking, and semantic segmentation, making event sensing increasingly practical for real-time robotic perception \cite{chakravarthi2024recent}.

While much of event vision has focused on estimating the current state of the scene, many robotics pipelines benefit from \emph{anticipation} \cite{mcallister2022control}. Short-horizon prediction can improve safety and stability by providing early evidence of motion and scene dynamics, rather than reacting after changes occur \cite{gehrig2024low}. This motivates \emph{event-based future prediction}: forecasting future event representations from a short history of event observations. Accurate future event predictions are valuable not only as an end task but also as a reusable intermediate signal that can support multiple downstream modules \cite{wu2024motion}. For instance, predicted future events can support object detection, semantic segmentation, improve short-term tracking consistency, and provide early perception cues for navigation and collision avoidance. In these settings, even modest improvements in short-horizon forecasting can translate into meaningful gains in robustness when future sensor readings are not yet available but actions must still be taken \cite{gehrig2024low, bhattacharya2024monocular}.

Despite its importance, practical event forecasting remains constrained by dominant modeling choices. Diffusion-style predictors can improve sample fidelity but often rely on iterative sampling and heavy pretrained backbones, increasing inference latency, memory cost, and engineering overhead \cite{wu2024motion,nichol2021improved}. Meanwhile, forecasting architectures developed for dense RGB frames are not tailored to sparse, polarity-structured event streams, and direct adaptation can bias learning toward background-dominant pixel fidelity rather than preserving motion-critical event structure. This motivates a lightweight, deterministic, single-pass event forecaster that preserves structure under tight latency and memory budgets.

In this work, we introduce \textbf{E-TIDE} (\underline{E}vent-\underline{T}emporal \underline{I}nteraction-\underline{D}riven n\underline{E}twork), a lightweight model for event-based motion forecasting. \textbf{E-TIDE} operates directly on polarity-separated event representations and performs deterministic, single-pass forecasting, avoiding multi-step generative sampling, iterative refinement, and complex alignment procedures (detailed in Sec.~\ref{sec:method}). The resulting design is simple to reproduce and suitable for real-time deployment under tight latency and memory budgets.

At the core of \textbf{E-TIDE} is \textbf{TIDE} (\underline{T}emporal \underline{I}nteraction for \underline{D}ynamic \underline{E}vents), a lightweight temporal module designed for polarity-separated event data. By packing time into channels and using fully-parallel interactions, \textbf{TIDE} avoids recurrent state updates and enables fast, predictable single-pass inference with a small memory footprint. In addition, we employ a \textbf{channel-aware weighted focal objective} with separate ON/OFF polarity weights to address the extreme imbalance in event data, allocating learning capacity to rare but motion-critical activations while remaining stable on the dominant inactive background.

\textbf{E-TIDE} achieves strong structural overlap on future events while being genuinely deployable: a full 10$\rightarrow$10 rollout runs in single-digit milliseconds with a fraction of the VRAM and parameters required by existing predictors, whether RGB-retrained or event-native (see Table~\ref{tab:main_etram}). Overall, E-TIDE combines structure-preserving predictions with highly efficient single-pass inference, making event forecasting practical under the strict latency and memory constraints of real-time robotic perception. We evaluate on two real-world event benchmarks spanning complementary regimes: traffic monitoring (\emph{eTraM}) and heterogeneous high-speed motion (\emph{E-3DTrack}) \cite{etram, li20243d}. \footnote{Code and trained models will be available at: \url{https://github.com/biswadeep20666/E-TIDE}.}

Our contributions are summarized as follows:
\begin{itemize}
    \item We propose \textbf{E-TIDE}, a lightweight end-to-end model for event-based motion forecasting that performs \emph{deterministic, single-pass} prediction, avoiding diffusion-style sampling, iterative refinement, and large-scale pretraining. To our knowledge, E-TIDE is the only event-based motion forecaster that is directly deployable in real time under tight latency and memory budgets.
    \item We introduce a \textbf{polarity-aware weighted focal objective} with separate ON/OFF weights, providing explicit control over the contribution of each polarity and a practical adaptation to extreme ON/OFF imbalance in event data. This improves learning of sparse, motion-critical activations while allowing flexible emphasis between ON and OFF events.

\end{itemize}

\section{Related Work}

\textbf{RGB Spatiotemporal Forecasting.} Most established spatiotemporal predictors are designed for dense 3-channel (RGB) frames and optimize appearance-centric objectives, which do not directly match the sparse, polarity-structured nature of event streams. Classical recurrent predictors such as ConvLSTM and PredRNN model dynamics via sequential hidden-state updates \cite{convlstm,predrnn}. Follow-up recurrent variants such as MAU and SwinLSTM extend the same sequential paradigm with motion-aware and transformer-enhanced updates to better capture complex dynamics \cite{chang2021mau,tang2023swinlstm}. However, recurrence remains inherently sequential and can become a latency bottleneck, motivating efficient feed-forward or attention-based alternatives such as SimVP and TAU \cite{simvp,tan2023tau}. Several RGB methods further incorporate explicit motion structure: MIMO-VP jointly predicts multiple future frames to reduce error accumulation \cite{ning2023mimo}, while MMVP formulates prediction using a motion-matrix representation and emphasizes a compact design with reduced parameter overhead \cite{zhong2023mmvp}. Motion-graph-based forecasting pushes this further, replacing dense convolutions with sparse graph operations to substantially reduce model size and memory at competitive accuracy \cite{zhong2024motion}. Wavelet-driven spatiotemporal learning (WaST) follows the same efficiency-oriented goal from a frequency-domain perspective, modeling dynamics via multi-scale wavelet decompositions to reduce spatiotemporal redundancy and enable faster forecasting while preserving fine details \cite{wavelet}. In contrast to these efficiency-oriented designs, recent large-scale RGB forecasting models push long-horizon quality via transformer backbones, memory augmentation, and often extensive pretraining; however, these gains typically come with heavy computation and nontrivial training/inference pipelines that are not end-to-end lightweight and are difficult to transfer to the smaller, domain-specific event-camera datasets typically available
 \cite{park2025videotitans, shrivastava2024cvp}. While these approaches perform well on RGB benchmarks, direct adaptation to event data is often suboptimal because dense-frame assumptions and pixel-fidelity training bias learning toward background dominance rather than preserving the sparse, motion-critical event structure.

\textbf{Event-based Motion Forecasting.} Explicit forecasting of future \emph{event representations} remains far less studied than RGB video prediction: most event-vision work targets state estimation (e.g., flow, detection, tracking, odometry) rather than predicting future event observations \cite{chakravarthi2024recent}. A notable early step is predictive event representation modeling aimed at energy-efficient sensing, where a lightweight predictor estimates the next event frame primarily to enable suppression/gating and reduce downstream computation, typically in a one-step setting rather than multi-frame rollouts \cite{bu2023predictive}. In contrast, \textbf{E-Motion} is, to our knowledge, the closest existing work that directly matches our problem formulation of multi-step future event-frame simulation from a short history, casting event forecasting as a diffusion-based generation problem with iterative denoising and large pretrained backbones \cite{wu2024motion}. While diffusion improves sample fidelity, its multi-step sampling and scale incur substantial inference latency, memory cost, and engineering overhead, which can conflict with the strict real-time constraints of robotic perception pipelines \cite{gallego2020event,shih2023parallel,nichol2021improved}. In summary, RGB predictors, whether efficiency-oriented or large-scale, are not tailored to sparse, polarity-structured event streams, while the closest multi-step event-frame predictor relies on iterative diffusion with substantial compute overhead. This leaves a clear niche for lightweight, deterministic event forecasting under strict latency and memory budgets, which \textbf{E-TIDE} addresses via efficient single-pass prediction of polarity-separated event occurrence maps that preserve motion-critical structure.

% --------------------------- Method ---------------------------
\section{Method}
\label{sec:method}

\begin{figure*}[t]
  \centering
  \includegraphics[width=\textwidth]{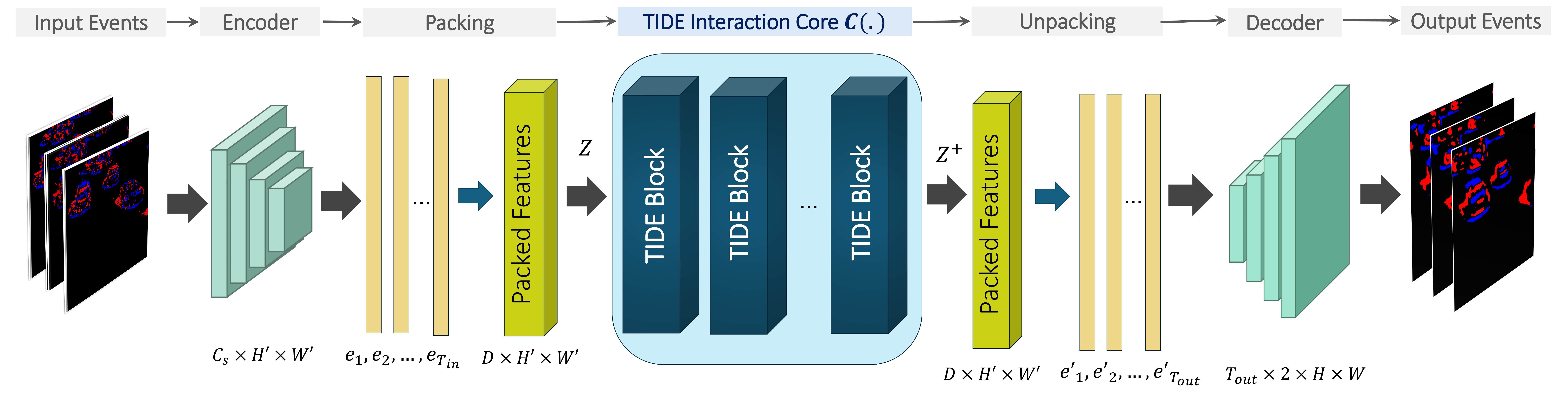}
  \caption{\textbf{E-TIDE overview.}
Polarity-separated event occurrence maps $\mathbf{X}=\{\mathbf{x}_t\}_{t=1}^{T_{\mathrm{in}}}$ are encoded with shared weights to obtain per-step features $\mathbf{e}_t$, which are packed along time into channels to form $\mathbf{Z}\in\mathbb{R}^{B\times D\times H'\times W'}$ with $D=T_{\mathrm{in}}C_s$ for fully-parallel spatiotemporal processing.
The interaction core $\mathcal{C}(\cdot)$ is a stack of \textbf{TIDE} blocks: within each block we denote the input as $\mathbf{U}$ and output as $\mathbf{U}^{+}$ (with $\mathbf{U}=\mathbf{Z}$ for the first block), and the stacked core maps $\mathbf{Z}$ to $\mathbf{Z}^{+}=\mathcal{C}(\mathbf{Z})$.
\emph{Unstack} denotes a reshape/view of the packed tensor (time-in-channels) for visualization; the decoder operates on $\mathbf{Z}^{+}$.
The decoder upsamples to future logits $\hat{\mathbf{S}}\in\mathbb{R}^{T_{\mathrm{out}}\times 2\times H\times W}$ and probabilities $\hat{\mathbf{Y}}=\sigma(\hat{\mathbf{S}})$.
We detail the internal \textbf{TIDE block} design in Fig.~\ref{fig:tide_module}.}
\label{fig:etide_overview}
\end{figure*}

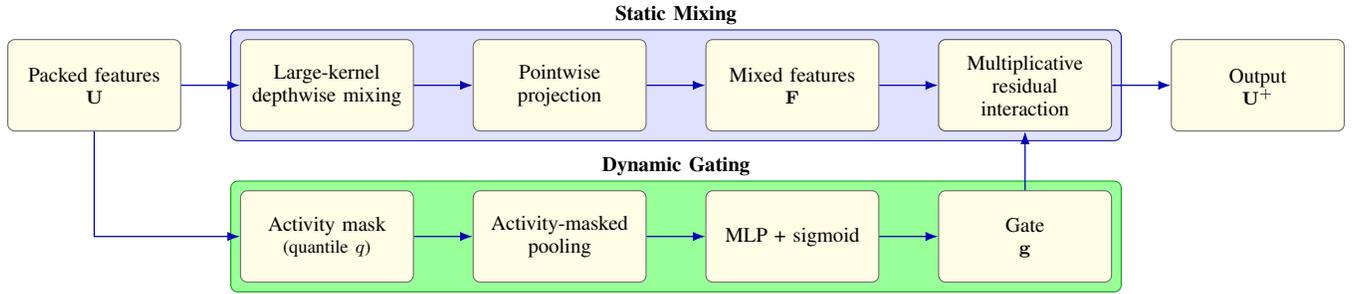
\begin{figure*}[t]
\centering
\resizebox{\textwidth}{!}{%
\begin{tikzpicture}[
  font=\small,
  box/.style={
    draw=black!60, rounded corners=3pt, align=center,
    minimum height=14mm, text width=24mm, inner sep=3.5pt,
    fill=yellow!12, line width=0.45pt
  },
  fusion/.style={
    draw=black!70, rounded corners=3pt, align=center,
    minimum height=14mm, text width=24mm, inner sep=3.5pt,
    fill=yellow!12, line width=0.45pt
  },
  arr/.style={-Latex, line width=0.55pt, draw=blue!70!black},
  node distance=7mm
]
% ═══════ INPUT ═══════
\node[box] (u) {Packed features\\[-0.5pt]{\footnotesize $\mathbf{U}$}};
% ═══════ TOP PATH ═══════
\node[box, right=9mm of u]   (lk) {Large-kernel\\depthwise mixing};
\node[box, right=9mm of lk]  (pw) {Pointwise\\projection};
\node[box, right=9mm of pw]  (f)  {Mixed features\\[-0.5pt]{\footnotesize $\mathbf{F}$}};
% ═══════ BOTTOM PATH ═══════
\node[box, below=9mm of lk]   (mask) {Activity mask\\[-0.5pt]{\footnotesize (quantile $q$)}};
\node[box, right=9mm of mask]  (pool) {Activity-masked\\pooling};
\node[box, right=9mm of pool]  (mlp)  {MLP + sigmoid};
\node[box, right=9mm of mlp] (g) {Gate\strut\\[-0.5pt]{\footnotesize $\mathbf{g}$}};
% ═══════ FUSION + OUTPUT ═══════
\node[fusion, right=9mm of f]  (mul) {Multiplicative\\residual\\interaction};
\node[box, right=9mm of mul] (out) {Output\strut\\[-0.5pt]{\footnotesize $\mathbf{U}^{+}$}};

% background lanes (add: \usetikzlibrary{fit,backgrounds})
\begin{scope}[on background layer]
\node[draw=blue!60!black,  fill=blue!12,  rounded corners=3pt, inner sep=4pt, fit=(lk)(pw)(f)(mul)] {};
\node[draw=green!60!black, fill=green!40, rounded corners=3pt, inner sep=4pt, fit=(mask)(pool)(mlp)(g)] {};
\end{scope}

\node[anchor=south, font=\bfseries\small]
  at ($(lk.north)!0.5!(mul.north) + (0,1mm)$) {Static Mixing};

\node[anchor=south, font=\bfseries\small]
  at ($(mask.north)!0.5!(g.north) + (0,1mm)$) {Dynamic Gating};
% ═══════ ARROWS — top path ═══════
\draw[arr] (u)  -- (lk);
\draw[arr] (lk) -- (pw);
\draw[arr] (pw) -- (f);
\draw[arr] (f)  -- (mul);
\draw[arr] (mul) -- (out);
% ═══════ ARROWS — gating branch ═══════
\draw[arr] (u)    |- (mask);
\draw[arr] (mask)  -- (pool);
\draw[arr] (pool)  -- (mlp);
\draw[arr] (mlp)   -- (g);
\draw[arr] (g.north) -- ++(0,4.5mm) -| (mul.south);
\end{tikzpicture}%
}
\caption{\textbf{TIDE block (used in Fig.~\ref{fig:etide_overview}).}
Given packed features $\mathbf{U}$, \textit{static mixing} performs large-kernel depthwise spatial aggregation followed by a pointwise projection to produce mixed features $\mathbf{F}$, while \textit{dynamic gating} computes an activity mask (quantile $q$), pools only active regions, and generates a channel gate $\mathbf{g}$ via an MLP and sigmoid. The two paths fuse through the multiplicative residual interaction $\mathbf{U}^{+}=\mathbf{g}\odot\mathbf{F}\odot(\mathbf{1}+\mathbf{U})$.}
\label{fig:tide_module}
\end{figure*}

\noindent\textbf{Overview.}
Given $T_{\mathrm{in}}$ polarity-separated event occurrence maps $\mathbf{X}$,
\textbf{E-TIDE} first extracts per-step spatial features with a shared encoder
and packs time into channels to form a single tensor
$\mathbf{Z}\in\mathbb{R}^{B\times (T_{\mathrm{in}}C_s)\times H'\times W'}$.
A fully-parallel \emph{interaction core} $\mathcal{C}(\cdot)$, implemented by
stacking $N_T$ \textbf{TIDE} blocks, performs spatiotemporal interaction on
$\mathbf{Z}$ using only 2D operators.
A decoder then maps the processed tensor back to future logits
$\hat{\mathbf{S}}\in\mathbb{R}^{T_{\mathrm{out}}\times 2\times H\times W}$
(and probabilities $\hat{\mathbf{Y}}=\sigma(\hat{\mathbf{S}})$).
Training uses a polarity-aware focal objective tailored to event sparsity,
optionally augmented with a temporal difference regularizer to encourage
realistic temporal evolution.

% ---------------------------------------------------------------
\subsection{Problem setup}
\label{sec:setup}
% ---------------------------------------------------------------

\subsubsection{Event occurrence maps}
An event camera produces an asynchronous stream of events
$e_i=(u_i,v_i,t_i,p_i)$, where $(u_i,v_i)$ is pixel location, $t_i$ is timestamp,
and $p_i\in\{+1,-1\}$ is the polarity.
We discretize time into bins $\mathcal{I}_t=[\tau_t,\tau_{t+1})$ and aggregate
events within each bin into a two-channel \emph{binary occurrence} map
$\mathbf{x}_t\in\{0,1\}^{2\times H\times W}$:
\begin{equation}
\begin{aligned}
x_t^{c}(u,v)
&= \mathbb{I}\!\Big[
\exists\, i:\ (u_i,v_i)=(u,v)\ \land\ p_i=c\ \land\ t_i\in\mathcal{I}_t
\Big],\\
&\qquad c\in\{+1,-1\},
\end{aligned}
\end{equation}
and we stack polarities as $\mathbf{x}_t=\big[x_t^{+};x_t^{-}\big]$.

\subsubsection{Forecasting objective}
Given an input sequence of occurrence maps
\begin{equation}
\mathbf{X}=\{\mathbf{x}_t\}_{t=1}^{T_{\mathrm{in}}},
\qquad \mathbf{x}_t \in \{0,1\}^{2 \times H \times W},
\end{equation}
our goal is to predict the future sequence
\begin{equation}
\mathbf{Y}=\{\mathbf{y}_t\}_{t=1}^{T_{\mathrm{out}}},
\qquad \mathbf{y}_t \in \{0,1\}^{2 \times H \times W}.
\end{equation}
The network outputs logits
$\hat{\mathbf{S}}\in\mathbb{R}^{T_{\mathrm{out}}\times 2\times H\times W}$
and probabilities $\hat{\mathbf{Y}}=\sigma(\hat{\mathbf{S}})$.

% ---------------------------------------------------------------
\subsection{Encoder--interaction--decoder architecture}
\label{sec:arch}
% ---------------------------------------------------------------
\textbf{E-TIDE} follows an encoder--interaction--decoder design
(Fig.~\ref{fig:etide_overview}):
\begin{equation}
  \hat{\mathbf{S}} = f_\theta(\mathbf{X})
  = \mathrm{Dec}\!\left(\mathcal{C}\!\left(\mathrm{Enc}(\mathbf{X})\right)\right),
  \qquad \hat{\mathbf{Y}}=\sigma(\hat{\mathbf{S}}),
\end{equation}
where $\mathcal{C}(\cdot)$ is a fully-parallel interaction core built from our
proposed \textbf{TIDE} modules
(\textbf{T}emporal \textbf{I}nteraction for \textbf{D}ynamic \textbf{E}vents) shown in Fig \ref{fig:tide_module}.

\paragraph{Encoder}
Each input frame is embedded with shared weights:
\begin{equation}
  \mathbf{e}_t = \mathrm{Enc}(\mathbf{x}_t),
  \qquad \mathbf{e}_t \in \mathbb{R}^{C_s \times H' \times W'},
\end{equation}
where $C_s$ is the number of feature channels per time step and $(H',W')$ is the
encoder output resolution.
In practice, we reshape
$\mathbf{X}\in\mathbb{R}^{B\times T_{\mathrm{in}}\times 2\times H\times W}$
into $\mathbb{R}^{(B\,T_{\mathrm{in}})\times 2\times H\times W}$ and apply a
stack of strided convolutions to obtain the latent features at resolution
$(H',W')$.

\paragraph{Time-to-channel packing}
To avoid sequential recurrence, we pack time into the channel dimension:
\begin{equation}
  \mathbf{Z} = \phi(\{\mathbf{e}_t\}_{t=1}^{T_{\mathrm{in}}})
  \in \mathbb{R}^{B\times D \times H' \times W'},
  \qquad D = T_{\mathrm{in}} C_s,
\end{equation}
where $\phi(\cdot)$ concatenates features along channels.
This enables fully-parallel spatiotemporal interaction with standard 2D
operators, following the general philosophy of efficient spatiotemporal
predictive blocks~\cite{tan2023tau}.

\paragraph{Decoder}
The decoder maps the processed tensor back to the output sequence and upsamples
to $(H,W)$:
\begin{equation}
  \hat{\mathbf{S}} = \mathrm{Dec}(\mathbf{Z}^+),
  \qquad \mathbf{Z}^+ = \mathcal{C}(\mathbf{Z}).
\end{equation}
Here
$\mathcal{C}\!:\mathbb{R}^{B\times D\times H'\times W'}
\!\rightarrow\! \mathbb{R}^{B\times D\times H'\times W'}$
denotes our interaction core, implemented as a stack of $N_T$ \textbf{TIDE}
modules (Sec.~\ref{sec:tide}), which transforms the packed tensor $\mathbf{Z}$
into $\mathbf{Z}^+ = \mathcal{C}(\mathbf{Z})$.

% ---------------------------------------------------------------
\subsection{TIDE Block}
\label{sec:tide}
% ---------------------------------------------------------------
We denote the packed tensor entering each TIDE block as $\mathbf{U}$
(with $\mathbf{U}=\mathbf{Z}$ at the first block).
In sparse event tensors most spatial locations are inactive, and motion evidence
concentrates in a small subset of pixels.
This sparsity makes standard global pooling dominated by background and additive
residual updates insensitive to rare activations.
\textbf{TIDE} addresses this with three components:
(i)~large-kernel depthwise spatial mixing to capture context efficiently,
(ii)~an activity-masked channel gate that focuses the summary signal on
high-activity regions, and
(iii)~a multiplicative residual $(1+\mathbf{U})$ that acts as a soft identity
on inactive locations while amplifying gradient flow where motion evidence
exists.

\paragraph{Large-kernel depthwise mixing}
Two depthwise convolution layers achieve large receptive fields at low cost: a
standard depthwise convolution with kernel size $k_1$, followed by a
\emph{dilated} depthwise convolution with kernel size $k_2$ and dilation rate
$\rho$.
A pointwise ($1\!\times\!1$) projection then mixes channels:
\begin{align}
  \mathbf{A}_1 &= \mathrm{DWConv}_{k_1}(\mathbf{U}), \\
  \mathbf{A}_2 &= \mathrm{DWConv}^{\mathrm{dil}(\rho)}_{k_2}(\mathbf{A}_1), \\
  \mathbf{F}   &= \mathrm{PWConv}(\mathbf{A}_2).
\end{align}

\paragraph{Activity-masked global pooling}
Global average pooling can be dominated by inactive regions.
We compute a per-location activity map
\begin{equation}
  \mathbf{m} = \frac{1}{D}\sum_{j=1}^{D} |\mathbf{U}_j|
  \;\in \mathbb{R}^{B\times 1\times H'\times W'},
\end{equation}
and retain only high-activity locations via a quantile threshold:
\begin{equation}
  \delta = \operatorname{Quantile}\,\bigl(\operatorname{vec}(\mathbf{m}),\, q\bigr),
  \qquad
  \mathbf{M} = \mathbb{I}\!\left[\mathbf{m} \ge \delta\right].
\end{equation}

Features are then pooled over this mask:
\begin{equation}
  \operatorname{Pool}(\mathbf{U}) =
  \frac{\sum_{h,w}\mathbf{M}_{h,w}\,\mathbf{U}_{:,:,h,w}}
       {\sum_{h,w}\mathbf{M}_{h,w}+\varepsilon}
  \;\in \mathbb{R}^{B\times D}.
\end{equation}

\paragraph{Channel gate and multiplicative residual interaction}
A lightweight MLP with reduction ratio $r$ produces a channel gate:
\begin{equation}
  \mathbf{g} = \sigma\!\Big(\mathbf{W}_2\,\rho\!\left(\mathbf{W}_1\,
  \mathrm{Pool}(\mathbf{U})\right)\!\Big)
  \;\in \mathbb{R}^{B\times D},
\end{equation}
where $\rho(\cdot)$ is ReLU and $\sigma(\cdot)$ is sigmoid.
After reshaping $\mathbf{g}$ to $B\!\times\!D\!\times\!1\!\times\!1$, the
module output is
\begin{equation}
  \mathrm{TIDE}(\mathbf{U})
  = \mathbf{g}\odot \mathbf{F}\odot(\mathbf{1}+\mathbf{U}).
  \label{eq:tide}
\end{equation}

\noindent\emph{Gradient perspective}\;
Treating $\mathbf{g}$ and $\mathbf{F}$ as locally fixed
(a straight-through approximation), the local Jacobian of
Eq.~\eqref{eq:tide} w.r.t.\ $\mathbf{U}$ scales as
$\mathbf{g}\odot\mathbf{F}$.
Inactive regions typically satisfy $\mathbf{F}\!\approx\!\mathbf{0}$ and
receive near-zero updates, whereas active regions yield larger
$\mathbf{F}$ and thus amplified gradient flow.
The bounded sigmoid gate $\mathbf{g}$ and layer normalization regulate the
overall gain.

\paragraph{Block form}
We use a standard pre-norm residual block with stochastic depth:
\begin{align}
  \mathbf{U}' &= \mathbf{U}
    + \mathrm{DropPath}\!\bigl(\mathrm{TIDE}(\mathrm{LN}(\mathbf{U}))\bigr),\\
  \mathbf{U}^{+} &= \mathbf{U}'
    + \mathrm{DropPath}\!\bigl(\mathrm{FFN}(\mathrm{LN}(\mathbf{U}'))\bigr),
\end{align}
where the FFN consists of two pointwise convolutions with GELU activation and
expansion ratio~$e$.
Stacking $N_T$ blocks yields the interaction core $\mathcal{C}(\cdot)$.
Implementation details including all hyperparameters
($k_1,k_2,\rho,r,q,N_T,e$) are given in Sec.~\ref{sec:experiments}.

% ---------------------------------------------------------------
\subsection{Training objective}
\label{sec:loss}
% ---------------------------------------------------------------
Event tensors exhibit severe class imbalance, and the two polarity channels
often differ markedly in sparsity.
Treating all pixels and both polarities uniformly can bias learning toward
trivial inactive predictions.
We address this with a \textbf{polarity-aware weighted focal objective} and a temporal difference regularizer.

\paragraph{Polarity-aware focal loss}
Let $\hat{s}_{t,c,h,w}$ denote a predicted logit and
$\hat{p}_{t,c,h,w}=\sigma(\hat{s}_{t,c,h,w})$.
For binary targets $y_{t,c,h,w}\!\in\!\{0,1\}$, the focal form is:
\begin{align}
  \mathcal{L}_{\mathrm{FL}}(\hat{p}, y) =\;
  &{-}\alpha\,y\,(1{-}\hat{p})^{\gamma}\log(\hat{p}+\varepsilon) \nonumber\\
  &{-}(1{-}\alpha)(1{-}y)\,\hat{p}^{\gamma}\log(1{-}\hat{p}+\varepsilon),
\end{align}
with focusing parameter $\gamma$ and balancing parameter $\alpha$.
We introduce per-channel weights $\lambda_{+},\lambda_{-}$
(normalized so that $\lambda_{+}+\lambda_{-}=1$) to reallocate gradient budget
across polarities:
% \begin{equation}
%   \mathcal{L}_{\mathrm{pol\text{-}FL}} =
%   \frac{1}{T_{\mathrm{out}}HW}
%   \sum_{t=1}^{T_{\mathrm{out}}}\sum_{h,w}
%   \sum_{c\in\{+,-\}}
%   \lambda_c \,
%   \mathcal{L}_{\mathrm{FL}}(\hat{p}_{t,c,h,w},\, y_{t,c,h,w}).
%   \label{eq:polfocal}
% \end{equation}

\begin{equation}
  \mathcal{L}_{\mathrm{pol\text{-}FL}}
  =
  \frac{1}{|\Omega|}
  \sum_{i\in\Omega}
  \sum_{c\in\{+,-\}}
  \lambda_c\,
  \mathcal{L}_{\mathrm{FL}}(\hat{p}_{i,c}, y_{i,c}),
  \label{eq:polfocal}
\end{equation}
where $\Omega=[T_{\mathrm{out}}]\times[H]\times[W]$, $[n]=\{1,\ldots,n\}$, and $i=(t,h,w)$.

\paragraph{Temporal difference regularization}
Matching frames alone constrains first-order structure but leaves higher-order temporal dynamics underdetermined. We therefore align the distribution of inter-frame differences between prediction and ground truth, imposing a second-order temporal prior that suppresses spurious oscillations and enforces coherent motion evolution~\cite{tan2023tau}.

Let $\hat{\mathbf{Y}}_t=\sigma(\hat{\mathbf{S}}_t)$ and
$\mathbf{y}_t\in\{0,1\}^{2\times H\times W}$ (cast to float when differencing).
Define
\begin{equation}
\Delta \hat{\mathbf{y}}_t=\operatorname{vec}(\hat{\mathbf{Y}}_{t+1}-\hat{\mathbf{Y}}_t),\qquad
\Delta \mathbf{y}_t=\operatorname{vec}(\mathbf{y}_{t+1}-\mathbf{y}_t).
\end{equation}
With temperature $\tau$,
\begin{equation}
\mathbf{p}_t=\operatorname{softmax}(\Delta \hat{\mathbf{y}}_t/\tau),\qquad
\mathbf{q}_t=\operatorname{softmax}(\Delta \mathbf{y}_t/\tau),
\end{equation}
and we minimize
\begin{equation}
\mathcal{L}_{\mathrm{DDR}}=
\frac{1}{T_{\mathrm{out}}-1}\sum_{t=1}^{T_{\mathrm{out}}-1}
\mathrm{KL}(\mathbf{p}_t\,\|\,\mathbf{q}_t).
\end{equation}

\paragraph{Final loss}
The complete training objective is:
\begin{equation}
  \mathcal{L}
  = \mathcal{L}_{\mathrm{pol\text{-}FL}}
  + \alpha_{\mathrm{DDR}}\,\mathcal{L}_{\mathrm{DDR}},
\end{equation}
where $\alpha_{\mathrm{DDR}}$ controls the regularizer strength.

% --------------------------- End Method --------------------------

% =============================
% (Recommended float control — put in PREAMBLE)
% =============================
% \usepackage{placeins}      % provides \FloatBarrier
% \usepackage{dblfloatfix}   % improves 2-col float placement (safer than stfloats)
% % If tables still feel tight:
% % \setlength{\textfloatsep}{8pt}
% % \setlength{\floatsep}{6pt}
% % \setlength{\intextsep}{6pt}

% -----------------------------
% Experiments
% -----------------------------
\section{Experiments}
\label{sec:experiments}

\subsection{Experimental setup}
\label{sec:exp_setup}

\paragraph{{Datasets}}
We evaluate on \textbf{ETram} ~\cite{etram} and \textbf{E-3DTrack}~\cite{li20243d} using their official train/val/test splits.
We select these benchmarks to cover two distinct event-vision regimes:
\textbf{ETram} captures long-duration, static-view traffic monitoring with diverse road users (e.g., cars, two-wheelers, pedestrians), while
\textbf{E-3DTrack} stresses high-speed, non-homogeneous object motion with a wide variety of trajectories.
This pairing provides a strong testbed for assessing both scene-level traffic dynamics and extreme-motion generalization.

To form inputs, following previous work, we convert the event stream into 30\,Hz time bins~\cite{etram} and represent each bin as a two-channel binary occurrence map (ON/OFF), where each pixel indicates whether at least one event fired within the bin.
Since ETram is high-resolution and high-temporal-rate, to keep computation tractable we follow E-Motion~\cite{wu2024motion} and randomly sample 512$\times$512 crops with non-trivial activity, then downsample to 128$\times$128.
For E-3DTrack, which is provided at a smaller spatial resolution, we directly downsample the full frame to 128$\times$128.
All methods use the same temporal protocol: $T_{\mathrm{in}}{=}10 \rightarrow T_{\mathrm{out}}{=}10$.

\paragraph{Baselines}
We compare against E-Motion~\cite{wu2024motion}, a recent event-native forecasting method, and a broad set of strong RGB video prediction models that we retrain on event occurrence maps for a fair, input-matched comparison.
Our RGB-derived baselines cover recurrent predictors and modern fully-parallel architectures published in recent years: PredRNN~\cite{predrnn}, SimVP~\cite{simvp},
MIMO-VP~\cite{ning2023mimo}, MMVP~\cite{zhong2023mmvp}, and MGVP~\cite{zhong2024motion}.
All baselines are retrained under identical preprocessing and the same $10\rightarrow10$ protocol.

\paragraph{Metrics}
We evaluate prediction quality using pixel-level fidelity metrics (MSE$\downarrow$, SSIM$\uparrow$, LPIPS$\downarrow$), as well as overlap metrics (mIoU (mean ON/OFF IoU), and aIoU (IoU on the polarity-agnostic occupancy mask formed by OR-ing both channels)). These overlap metrics also indicate downstream usability, as they reflect whether predicted event structure is suitable for segmentation-style evaluation.
We apply a sigmoid to the predicted logits and binarize with a threshold $\tau$, then compute IoU separately for ON and OFF channels against ground-truth occurrence maps. We report mIoU (mean of ON/OFF IoU) and aIoU. To avoid manual tuning, $\tau$ is selected automatically per frame via a thresholding rule that maximizes inter-class variance~\cite{yang2020improved}, and overlap metrics are accumulated globally over the test set. As a robustness check, we also evaluate IoU across a grid of fixed thresholds and confirm that relative rankings remain unchanged, indicating that automatic thresholding does not advantage any particular method.

\paragraph{Implementation details} All methods use $T_{\mathrm{in}}{=}10 \rightarrow T_{\mathrm{out}}{=}10$ and are trained on one \textbf{NVIDIA RTX 3090} with batch size $4$. E-TIDE uses $N_T{=}4$, $C_s{=}8$, $D{=}T_{\mathrm{in}}C_s$, $k{=}3$, DropPath $0.2$, and TIDE parameters $(k_1,k_2){=}(5,7)$, $\rho{=}3$, $q{=}0.98$, $r{=}16$. We train with PAFL/DDR using $\alpha{=}0.75$, $\gamma{=}2.0$, $(\lambda_+,\lambda_-){=}(0.65,0.35)$, and $\alpha_{\mathrm{DDR}}{=}0.1$.

% ============================================================
% Results: ETram
% ============================================================
\subsection{Main results}
\label{sec:exp_etram}

\begin{figure*}[!t]
  \centering
  \includegraphics[width=\textwidth]{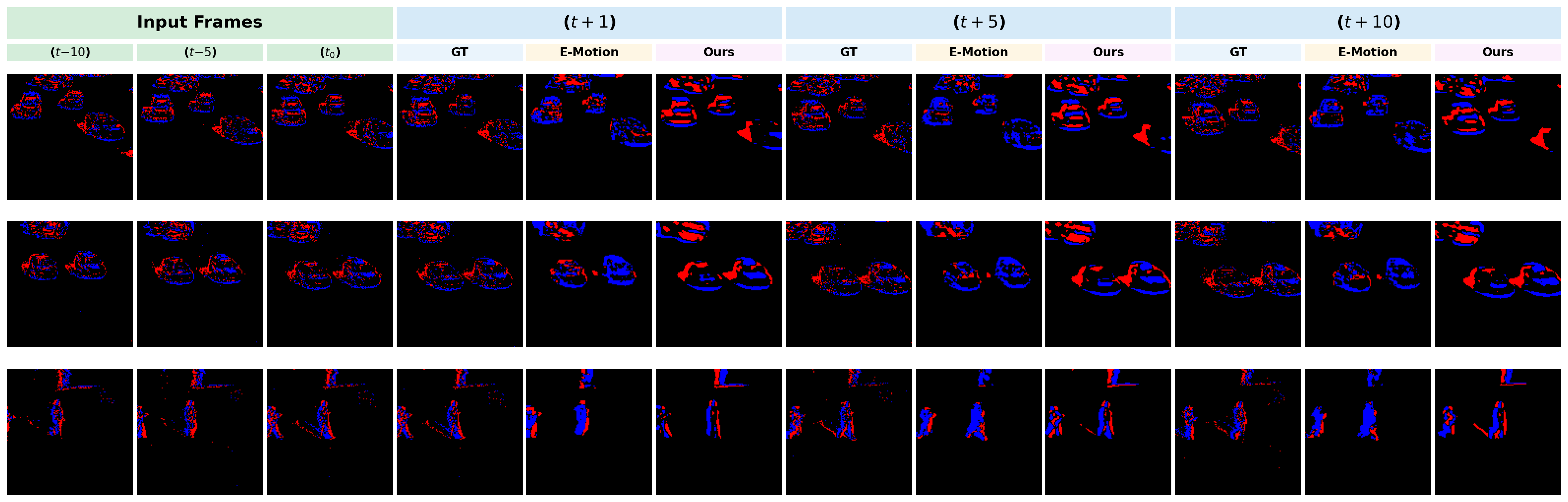}
  \caption{\textbf{Qualitative comparison on \textsc{eTraM}.}
Example $10\!\rightarrow\!10$ forecasts showing input event frames and future predictions at $t{+}1$, $t{+}5$, and $t{+}10$.
E-TIDE produces motion-aligned, temporally consistent event structure that better preserves thin boundaries compared to the event-native baseline E-Motion. Red/blue indicate ON/OFF polarities.}
  \label{fig:qual}
\end{figure*}

\begin{table*}[!t]
\centering
\caption{\textbf{ETram comparison} on $10\!\rightarrow\!10$ event prediction (batch size 1).
Lower is better for MSE/LPIPS; higher is better for SSIM/mIoU/aIoU.
Peak VRAM reports \texttt{max\_memory\_allocated} during inference on RTX~3090.
\textbf{Type:} \textbf{RGB} indicates RGB-origin video predictors retrained on event occurrence maps;
\textbf{Event} indicates event-native predictors designed for event data.}
\label{tab:main_etram}
\footnotesize
\setlength{\tabcolsep}{3.5pt}
\renewcommand{\arraystretch}{1.10}
\resizebox{\textwidth}{!}{%
\begin{tabular}{|l|c|c|c|c|c|c|c|c|c|c|}
\hline
\textbf{Method} & \textbf{Venue} & \textbf{Type} & \textbf{Params (M)} & \textbf{Time (ms)} & \textbf{VRAM (GB)} &
\textbf{MSE}$\downarrow$ & \textbf{SSIM}$\uparrow$ & \textbf{LPIPS}$\downarrow$ &
\textbf{mIoU}$\uparrow$ & \textbf{aIoU}$\uparrow$ \\
\hline
PredRNN \cite{predrnn} & \textcolor{teal}{NeurIPS'17} & RGB & 14.0 & 116.3 & 0.19 & 0.0416 & 0.773 & 0.324 & 0.515 & 0.563 \\
\hline
SimVP \cite{simvp} & \textcolor{teal}{CVPR'22} & RGB & 46.0 & \underline{10.6} & 0.68 & 0.0370 & 0.788 & \underline{0.360} & \underline{0.534} & \underline{0.571} \\
\hline
MIMO-VP \cite{ning2023mimo} & \textcolor{teal}{AAAI'23} & RGB & 22.6 & 56.16 & 1.27 & \textbf{0.0250} & \textbf{0.846} & 0.350 & 0.470 & 0.545 \\
\hline
MMVP \cite{zhong2023mmvp} & \textcolor{teal}{ICCV'23} & RGB & 14.8 & 127.7 & 9.65 & 0.0410 & 0.782 & 0.374 & 0.480 & 0.550 \\
\hline
MGVP \cite{zhong2024motion} & \textcolor{teal}{NeurIPS'24} & RGB & 0.16 & 252.7 & 0.15 & 0.0450 & 0.772 & 0.407 & 0.357 & 0.412 \\
\hline
E-Motion \cite{wu2024motion} & \textcolor{teal}{NeurIPS'24} & Event & 1521 & 8522 & 9.24 & 0.0462 & 0.744 & 0.387 & 0.362 & 0.423 \\
\hline
\textbf{E-TIDE} & \textbf{Ours} & \textbf{Event} & \textbf{0.4} & \textbf{3.1} & \textbf{0.12} &
\underline{0.0350} & \underline{0.800} & \textbf{0.325} & \textbf{0.551} & \textbf{0.601} \\
\hline
\end{tabular}}
\end{table*}

On \textsc{ETram} (Table~\ref{tab:main_etram}), E-TIDE achieves the strongest structural overlap, with
\textbf{mIoU/aIoU = 0.551/0.601}.
This substantially improves over E-Motion (\textbf{0.362/0.423}), yielding gains of \textbf{+0.189 mIoU} and \textbf{+0.178 aIoU}, while also improving
\textbf{SSIM} (0.800 vs 0.744) and \textbf{LPIPS} (0.325 vs 0.387).
Compared to strong RGB-origin predictors retrained on event occurence maps, E-TIDE improves over SimVP (\textbf{0.534/0.571}) by \textbf{+0.017 mIoU} and \textbf{+0.030 aIoU},
indicating better preservation of sparse, motion-aligned event structure in the multi-actor traffic setting.

On \textsc{E-3DTrack} (Table~\ref{tab:main_e3dtrack}), E-TIDE attains the best \textbf{aIoU = 0.6258} and improves over E-Motion (\textbf{0.5371}) by \textbf{+0.0887 aIoU}, while also achieving higher \textbf{SSIM} (0.9291 vs 0.8958).
E-TIDE additionally achieves the best \textbf{LPIPS} (0.1273), and remains competitive on \textbf{mIoU = 0.5234} (best 0.5477), supporting a favorable balance between overlap quality and perceptual consistency under high-speed, diverse motion patterns.
Relative to SimVP, E-TIDE also improves \textbf{aIoU} (0.6258 vs 0.6200) while maintaining comparable \textbf{SSIM} (0.9291 vs 0.9301). Although MIMO-VP leads dense MSE/SSIM on both datasets, these full-frame metrics can be dominated by inactive background regions under extreme event sparsity; we therefore emphasize mIoU/aIoU and downstream performance as stronger indicators of preserved motion structure.

\begin{figure*}[!t]
  \centering
  \includegraphics[width=\textwidth]{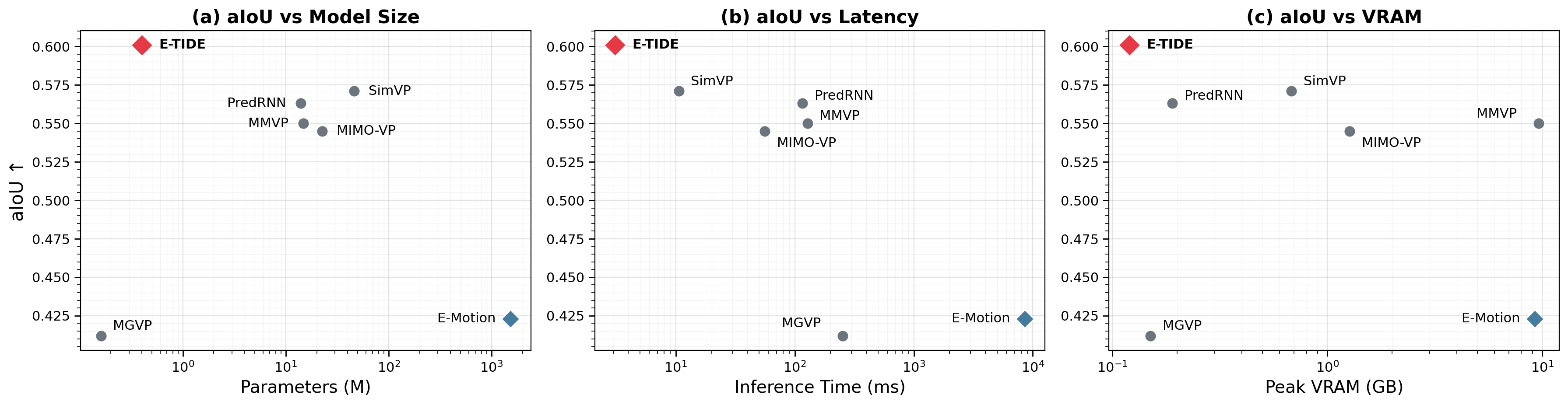}
  \caption{\textbf{Accuracy--efficiency Pareto trade-offs.}
  We plot aIoU (higher is better) against \textbf{model size}, \textbf{inference latency}, and \textbf{peak VRAM} (log-scale where indicated).
  Across all 3 plots, E-TIDE lies in the top-left frontier, ie.\ it beats the other models in size, latency and memory usage.}
  \label{fig:pareto_all}
\end{figure*}

With event occurrence maps formed at \textbf{30 Hz}, the $10\rightarrow10$ protocol forecasts 10 future occurrence maps, corresponding to a \textbf{333 ms} horizon. E-TIDE completes the full rollout in \textbf{3.1 ms} --- under 1\% of this horizon --- supporting real-time deployment in this forecasting setting, and far faster than E-Motion. Compared with E-Motion, E-TIDE uses $\mathbf{3800\times}$ fewer parameters, runs $\mathbf{2749\times}$ faster, and requires $\mathbf{77\times}$ less VRAM; compared with the strongest RGB-origin baseline SimVP, it uses $\mathbf{115\times}$ fewer parameters, runs $\mathbf{3.4\times}$ faster, and requires $\mathbf{5.7\times}$ less VRAM. Importantly, these reductions are achieved while improving \textbf{aIoU} on both datasets and improving \textbf{SSIM} on ETram, placing E-TIDE consistently on the accuracy--efficiency frontier (Fig \ref{fig:pareto_all}).

% ============================================================
% Results: E-3DTrack
% ===========================================================
\begin{table*}[!t]
\centering
\caption{\textbf{E-3DTrack comparison} on $10\!\rightarrow\!10$ event prediction (batch size 1).
Same metrics and measurement protocol as Table~\ref{tab:main_etram}.}
\label{tab:main_e3dtrack}
\footnotesize
\setlength{\tabcolsep}{3.5pt}
\renewcommand{\arraystretch}{1.10}
\resizebox{\textwidth}{!}{%
\begin{tabular}{|l|c|c|c|c|c|c|c|c|c|c|}
\hline
\textbf{Method} & \textbf{Venue} & \textbf{Type} & \textbf{Params (M)} & \textbf{Time (ms)} & \textbf{VRAM (GB)} &
\textbf{MSE}$\downarrow$ & \textbf{SSIM}$\uparrow$ & \textbf{LPIPS}$\downarrow$ &
\textbf{mIoU}$\uparrow$ & \textbf{aIoU}$\uparrow$ \\
\hline
PredRNN \cite{predrnn} & \textcolor{teal}{NeurIPS'17} & RGB & 14.0 & 116.3 & 0.19 & 0.013 & 0.916 & 0.212 & 0.416 & 0.526 \\
\hline
SimVP \cite{simvp} & \textcolor{teal}{CVPR'22} & RGB & 46.0 & \underline{10.6} & 0.68 & 0.0130 & \underline{0.9301} & \underline{0.1291} & 0.5208 & \underline{0.6200} \\
\hline
MIMO-VP \cite{ning2023mimo} & \textcolor{teal}{AAAI'23} & RGB & 22.6 & 56.16 & 1.27 & \textbf{0.0084} & \textbf{0.9416} & 0.1739 & \underline{0.5454} & 0.5824 \\
\hline
MMVP \cite{zhong2023mmvp} & \textcolor{teal}{ICCV'23} & RGB & 14.8 & 127.7 & 9.65 & 0.0324 & 0.7990 & 0.3347 & \textbf{0.5477} & 0.5976 \\
\hline
MGVP \cite{zhong2024motion} & \textcolor{teal}{NeurIPS'24} & RGB & 0.16 & 252.7 & 0.15 & 0.0152 & 0.9110 & 0.2303 & 0.3576 & 0.4457 \\
\hline
E-Motion \cite{wu2024motion} & \textcolor{teal}{NeurIPS'24} & Event & 1521 & 8522 & 9.24 & 0.0205 & 0.8958 & 0.1617 & 0.4189 & 0.5371 \\
\hline
\textbf{E-TIDE} & \textbf{Ours} & \textbf{Event} & \textbf{0.4} & \textbf{3.1} & \textbf{0.12} & \underline{0.0123} & 0.9291 & \textbf{0.1273} & 0.5234 & \textbf{0.6258} \\
\hline
\end{tabular}}
\end{table*}

% ============================================================
% Efficiency figure (placed AFTER both main tables)
% ============================================================

% ============================================================
% Downstream utility
% ============================================================
\subsection{Downstream utility: Segmentation and Tracking}
\label{sec:exp_downstream}

\begin{table}[t]
\centering
\footnotesize
\setlength{\tabcolsep}{4.5pt}
\renewcommand{\arraystretch}{1.12}
\caption{\textbf{Downstream evaluation using predicted events}}
\label{tab:downstream}
\begin{tabular}{lccccc}
\toprule
& \multicolumn{2}{c}{\textsc{ETram} Segmentation} & \multicolumn{3}{c}{\textsc{E-3DTrack} Tracking} \\
\cmidrule(lr){2-3}\cmidrule(lr){4-6}
Method & mIoU$\uparrow$ & aIoU$\uparrow$ & IDF1$\uparrow$ & IDP$\uparrow$ & IDR$\uparrow$ \\
\midrule
MMVP            & 0.457 & 0.491 & 0.7034 & 0.6954 & 0.6822 \\
MIMO-VP         & 0.459 & 0.483 & 0.7117 & 0.7161 & 0.7074 \\
E-Motion        & 0.432 & 0.441 & 0.6612 & 0.6732 & 0.6645 \\
\textbf{E-TIDE (Ours)} & \textbf{0.501} & \textbf{0.527} & \textbf{0.7605} & \textbf{0.8023} & \textbf{0.7227} \\
\bottomrule
\end{tabular}
\end{table}

\begin{figure*}[!t]
  \centering
  \includegraphics[width=\textwidth]{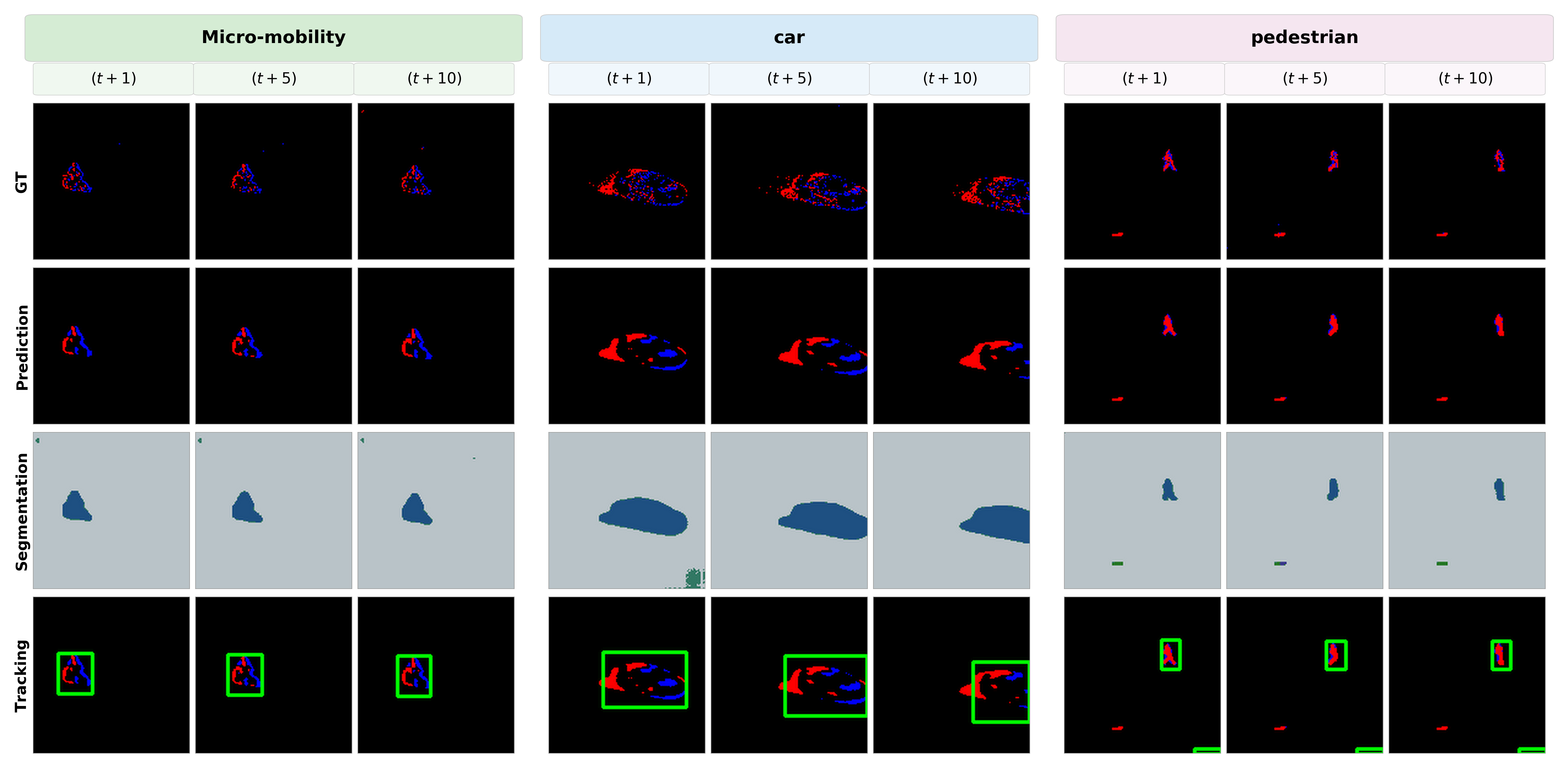}
  \caption{\textbf{Downstream tasks}
 Qualitative results for micro-mobility, cars, and pedestrians. Predicted event frames remain structurally faithful to the ground truth, enabling reliable segmentation and tracking across diverse object categories and forecast horizons.}
  \label{fig:downstream_tasks}
\end{figure*}

Beyond overlap metrics, we evaluate whether predicted futures are useful for downstream (Fig \ref{fig:downstream_tasks}).
Specifically, we use 10-step event forecasts as inputs to a standard, state-of-the-art segmentation pipeline\cite{zhang2023faster} on \textsc{ETram}, and to a tracking pipeline on \textsc{E-3DTrack}, which features high-speed, non-homogeneous object motion.
This setting tests whether a forecaster preserves motion-aligned, temporally consistent support that benefits boundary prediction and association, rather than producing fragmented activations that may score well under pixel-level fidelity metrics.

As shown in Table~\ref{tab:downstream}, E-TIDE yields the strongest downstream performance across both tasks, improving segmentation overlap on \textsc{ETram} and achieving the best identity-based tracking scores on \textsc{E-3DTrack}.
These results indicate that E-TIDE forecasts retain structured event evidence that is directly exploitable by subsequent perception modules, supporting its role as a lightweight predictive front-end for real-time pipelines.

% ============================================================
% Qualitative
% ===========================================================

\subsection{Qualitative results}
\label{sec:qual_results}

% Fig.~\ref{fig:qual} shows $10 \rightarrow 10$ forecasts on eTraM across increasing prediction horizons. Compared with E-Motion, E-TIDE produces temporally stable, motion-aligned event traces that better preserve thin structures, object contours, and inter-object separation. This difference becomes more visible at longer horizons: E-Motion tends to produce fragmented activations or spurious blobs, whereas E-TIDE maintains coherent motion trajectories and sparsity patterns that remain consistent with the ground truth. These qualitative results support the quantitative trends in Table~\ref{tab:main_etram}, where E-TIDE achieves stronger structural overlap while remaining substantially more efficient.

Fig.~\ref{fig:qual} shows $10\rightarrow10$ forecasts on eTraM across increasing prediction horizons.
Compared to E-Motion, E-TIDE produces temporally stable, motion-aligned event traces that better preserve thin structures, object contours, and inter-object separation. Notably, as the prediction horizon increases, E-Motion outputs fragmented or spurious blobs, whereas E-TIDE maintains coherent motion trajectories with sparsity patterns consistent with ground truth. This also corroborates the data in Table~\ref{tab:main_etram}.

% \begin{table}[!t]
% \centering
% \small
% \setlength{\tabcolsep}{6pt}
% \renewcommand{\arraystretch}{1.08}
% \caption{\textbf{E-TIDE ablations on \textsc{ETram}} ($10\rightarrow10$).}
% \label{tab:ablation_etide}
% \vspace{-3pt}
% \begin{tabular}{@{}lcc@{}}
% \hline
% \textbf{Variant} & \textbf{mIoU}$\uparrow$ & \textbf{aIoU}$\uparrow$ \\
% \hline
% E-TIDE w/o PAFL & 0.529 & 0.567 \\
% E-TIDE w/o activation-masked pooling & 0.539 & 0.578 \\
% E-TIDE w/o multiplicative residual & 0.532 & 0.571 \\
% E-TIDE w/o DDR & 0.542 & 0.589 \\
% \hline
% E-TIDE (full) & \textbf{0.551} & \textbf{0.601} \\
% \hline
% \end{tabular}
% \vspace{-10pt}
% \end{table}

\begin{table}[!t]
\centering
\small
\setlength{\tabcolsep}{6pt}
\renewcommand{\arraystretch}{1.08}
\caption{\textbf{E-TIDE ablations on \textsc{ETram}} ($10\rightarrow10$).}
\label{tab:ablation_etide}
\vspace{-3pt}
\begin{tabular*}{0.95\columnwidth}{@{\extracolsep{\fill}}lcc@{}}
\hline
\textbf{Variant} & \textbf{mIoU}$\uparrow$ & \textbf{aIoU}$\uparrow$ \\
\hline
E-TIDE w/o PAFL & 0.529 & 0.567 \\
E-TIDE w/o activation-masked pooling & 0.539 & 0.578 \\
E-TIDE w/o multiplicative residual & 0.532 & 0.571 \\
E-TIDE w/o DDR & 0.542 & 0.589 \\
\hline
E-TIDE (full) & \textbf{0.551} & \textbf{0.601} \\
\hline
\end{tabular*}
\vspace{-10pt}
\end{table}

Table~\ref{tab:ablation_etide} ablates key components of E-TIDE under a fixed backbone and training schedule.
Disabling any component reduces overlap, showing that performance is driven by complementary contributions.
The polarity-aware focal loss (PAFL) provides a strong gain over a non-polarity baseline, activation-masked pooling and the multiplicative residual further improve structure preservation, and the DDR term adds an additional boost, yielding the best overall mIoU/aIoU for the full model.

\section{Conclusion}

\begin{figure}[t]
\centering
\includegraphics[width=0.95\linewidth]{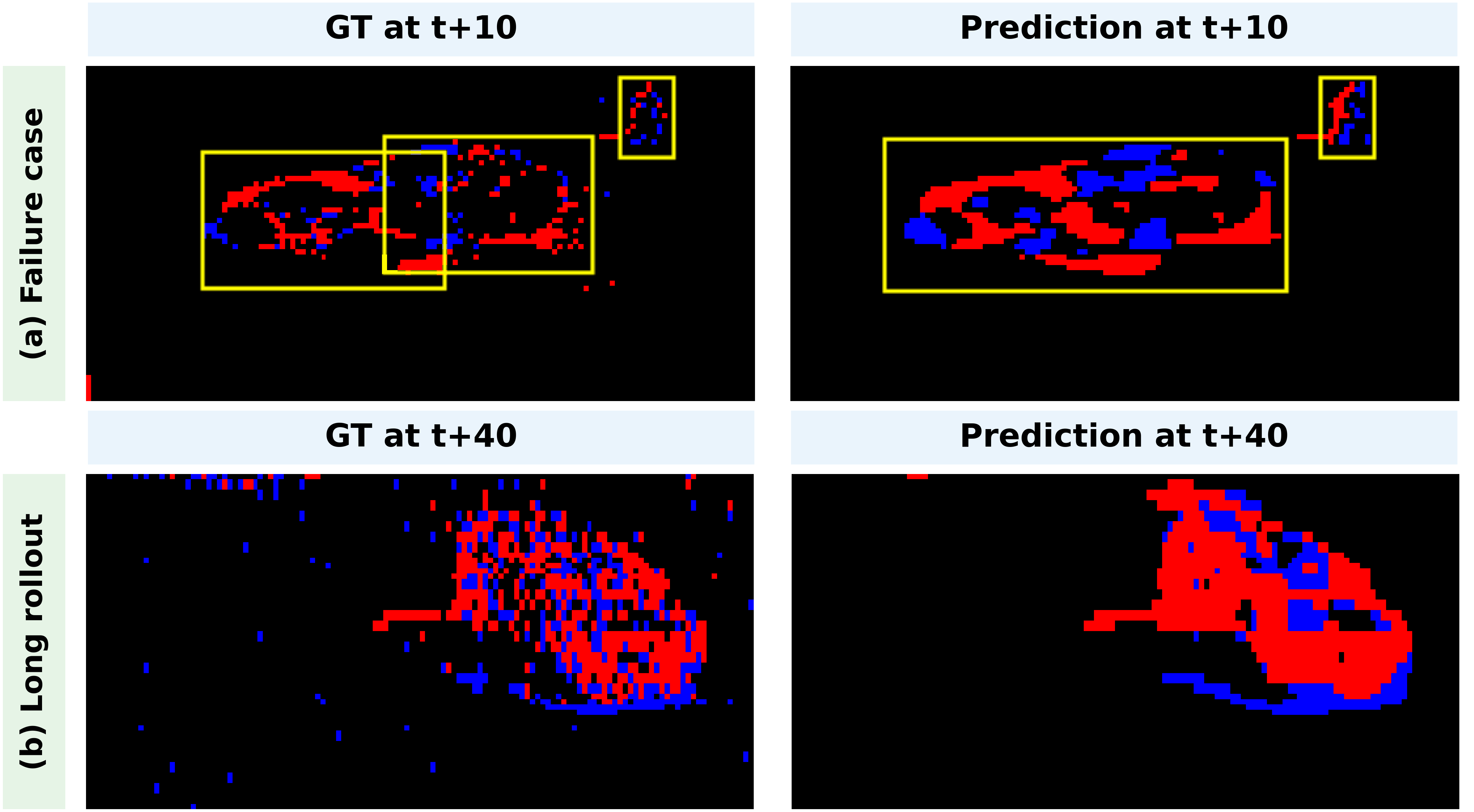}
\vspace{-4pt}
\caption{\textbf{Failure and degradation modes.} (a) Closely trailing objects may be merged under the standard $10\rightarrow10$ setting. (b) Extended recursive rollout at $t+40$ preserves main structure but smooths out fine event details.}
\label{fig:failure_cases}
\vspace{-6pt}
\end{figure}

We introduced \textbf{E-TIDE}, a lightweight and practical model for event-based future prediction on two-channel occurrence maps. By focusing on structure-preserving evaluation (mIoU/aIoU) rather than relying primarily on pixel-fidelity metrics that can be dominated by background agreement, our experiments highlight that accurate forecasting of sparse event support is achievable with a compact architecture. On ETram under a consistent 10$\rightarrow$10 protocol, \textbf{E-TIDE} attains the best overlap scores (mIoU 0.551, aIoU 0.601) while remaining highly efficient, using only 0.4M parameters, 0.12 GB peak VRAM, and 3.1 ms inference latency on an NVIDIA RTX 3090. These results make event prediction feasible under tight real-time and memory constraints, strengthening its suitability as a deployable predictive perception module. One limitation of our work is that closely overlapping instances can occasionally be merged (Fig.~\ref{fig:failure_cases}a). Under extended recursive rollouts, predictions remain spatially coherent at $t+40$, but can become over-smoothed and lose fine event structure (Fig.~\ref{fig:failure_cases}b).

Looking forward, E-TIDE can be further strengthened by more event-native computation. SNN-based temporal cores could reduce redundant dense computation and support neuromorphic low-latency deployment~\cite{rathi2023exploring}, while sparse or event-driven operators could improve scalability to higher resolutions and longer horizons. Extending E-TIDE beyond fixed-camera monitoring to moving-camera settings such as driving and aerial robotics is another promising direction, where ego-motion introduces new challenges for event-native forecasting.

% Looking forward, we see two promising directions to further improve efficiency and event-specific inductive bias. First, integrating spiking neural network (SNN) temporal cores could provide a more natural way to process event streams and reduce redundant dense computation\cite{eshraghian2023training}. Beyond efficiency, SNN-based designs also move toward neuromorphic deployment, enabling ultra-low-latency and energy-efficient implementations on dedicated spiking hardware \cite{rathi2023exploring}. Second, exploiting sparsity explicitly, via sparse convolutions, masked operators, or event-driven computation, may further cut unnecessary processing and enable scaling to higher resolutions and longer horizons without increasing latency or VRAM. Finally, while we focus here on fixed-camera forecasting, a core deployment regime for smart-city traffic monitoring and infrastructure sensing \cite{etram}, we are excited to extend \textbf{E-TIDE} to moving-camera settings (e.g., driving and aerial robotics), where ego-motion and dynamic viewpoints pose new challenges and opportunities for event-native forecasting.

% \begin{figure}[t]
%   \centering
%   \includegraphics[width=\columnwidth]{figures/failure_case.png}
%   \caption{\textbf{Failure case.} GT (left) vs prediction (right) at $t{+}10$: two closely trailing vehicles in GT are merged into one in the forecast.}
%   \label{fig:failure_case}
% \end{figure}

%NC: SNN also allow to go toward neuromorphic implementation for faster running. - FIXED!

\section*{ACKNOWLEDGMENT}

The authors thank the anonymous reviewers for constructive feedback. The computational work involved in this research was partially supported by NUS IT’s Research Computing group under grant NUSREC-HPC-00001. ChatGPT was used only for language polishing, grammar correction, and LaTeX formatting; all technical content, results, figures, and conclusions were developed and verified by the authors.

\bibliographystyle{IEEEtran}
\bibliography{bibilo}

\end{document}